\documentclass[sigconf]{acmart}

\usepackage{multirow}  
\usepackage[normalem]{ulem}

\usepackage{enumitem} 
\usepackage{siunitx}
\AtBeginDocument{%
  }

\copyrightyear{2026}
\acmYear{2026}
\setcopyright{cc}
\setcctype{by-nc-nd}
\acmConference[WWW '26]{Proceedings of the ACM Web Conference 2026}{April 13--17, 2026}{Dubai, United Arab Emirates}
\acmBooktitle{Proceedings of the ACM Web Conference 2026 (WWW '26), April 13--17, 2026, Dubai, United Arab Emirates}
\acmPrice{}
\acmDOI{10.1145/3774904.3792730}
\acmISBN{979-8-4007-2307-0/2026/04}

\begin{document}

\title[RPO-RAG: Aligning Small LLMs with Relation-aware Preference Optimization for KGQA]{RPO-RAG: Aligning Small LLMs with Relation-aware Preference Optimization for Knowledge Graph Question Answering}


\author{Kaehyun Um}
\orcid{0009-0005-2366-5296} 
\affiliation{%
\department{Department of Computer Science}
  \institution{Yonsei University}
  \city{Seoul}
  \country{Republic of Korea}
}
\email{khyun33@yonsei.ac.kr}

\author{KyuHwan Yeom}
\orcid{0009-0003-4252-6203} 
\affiliation{%
 \department{Department of Computer Science}
  \institution{Yonsei University}
  \city{Seoul}
  \country{Republic of Korea}
}
\email{tomma1121@yonsei.ac.kr}

\author{Haerim Yang}
\orcid{0009-0005-4664-6880} 
\affiliation{%
\department{Department of Artificial Intelligence}
  \institution{Yonsei University}
  \city{Seoul}
  \country{Republic of Korea}
}
\email{hly1013@yonsei.ac.kr}

\author{Minyoung Choi}
\orcid{0009-0007-3838-706X} 
\affiliation{%
\department{Department of Computer Science}
  \institution{Yonsei University}
  \city{Seoul}
  \country{Republic of Korea}
}
\email{min02choi@yonsei.ac.kr}

\author{Hyeongjun Yang}
\orcid{0000-0001-7958-224X} 
\affiliation{%
\department{Department of Computer Science}
  \institution{Yonsei University}
  \city{Seoul}
  \country{Republic of Korea}
}
\email{edbm95@yonsei.ac.kr}

\author{Kyong-Ho Lee}
\authornote{Corresponding author.}
\orcid{0000-0002-1581-917X} 
\affiliation{%
\department{Department of Computer Science}
  \institution{Yonsei University}
  \city{Seoul}
  \country{Republic of Korea}
}
\email{khlee89@yonsei.ac.kr}


\begin{abstract}
Large Language Models (LLMs) have recently demonstrated remarkable reasoning abilities, yet hallucinate on knowledge-intensive tasks. Retrieval-augmented generation (RAG) mitigates this issue by grounding answers in external sources, e.g., knowledge graphs (KGs). However, existing KG-based RAG approaches rely on semantics-unaware path sampling and are weakly aligned with KG reasoning objectives, which limits further accuracy gains. They also feed retrieved paths directly into the reasoner without organizing them into answer-centered reasoning paths, hindering small LLMs' ability to leverage the retrieved knowledge. Furthermore, prior works predominantly rely on large LLMs (e.g., ChatGPT/GPT-4) or assume backbones above 7B parameters, leaving sub-7B models underexplored.
We address this gap with \textbf{RPO-RAG}, the first KG-based RAG framework specifically designed for small LLMs, to the best of our knowledge. RPO-RAG introduces three key innovations: (1) a query-path semantic sampling strategy that provides informative supervisory signals; (2) a relation-aware preference optimization that aligns training with intermediate KG reasoning signals (e.g., relation); and (3) an answer-centered prompt design that organizes entities and reasoning paths in an interpretable format. 
Extensive experiments on two benchmark Knowledge Graph Question Answering (KGQA) datasets, WebQSP and CWQ, demonstrate that RPO-RAG effectively bridges the performance gap between small and large language models. On WebQSP, it improves F1 by up to 8.8\%, reflecting enhanced answer precision, while on CWQ it achieves new state-of-the-art results among models under 8B parameters in both Hit and F1. Overall, RPO-RAG substantially improves the reasoning capability of small LLMs—even under 3B parameters—highlighting their potential for resource-efficient and practical on-device KGQA applications.
\end{abstract}




\maketitle
\newcommand\webconfcodeurl{https://doi.org/10.5281/zenodo.18322650}
\newcommand\webconfmodelurl{https://doi.org/10.5281/zenodo.18322931}

\ifdefempty{\webconfcodeurl}{}{
\begingroup\small\noindent\raggedright\textbf{Resource Availability:}\\
The source code of this paper is publicly available at
\url{https://github.com/KaeHyun/RPO-RAG} and archived at
\url{\webconfcodeurl}.
The trained RPO-RAG models are publicly available and archived at
\url{\webconfmodelurl}.
\endgroup
}



\begin{figure}[h]
  \centering
  \includegraphics[width=1.2\linewidth, height=8cm, keepaspectratio]{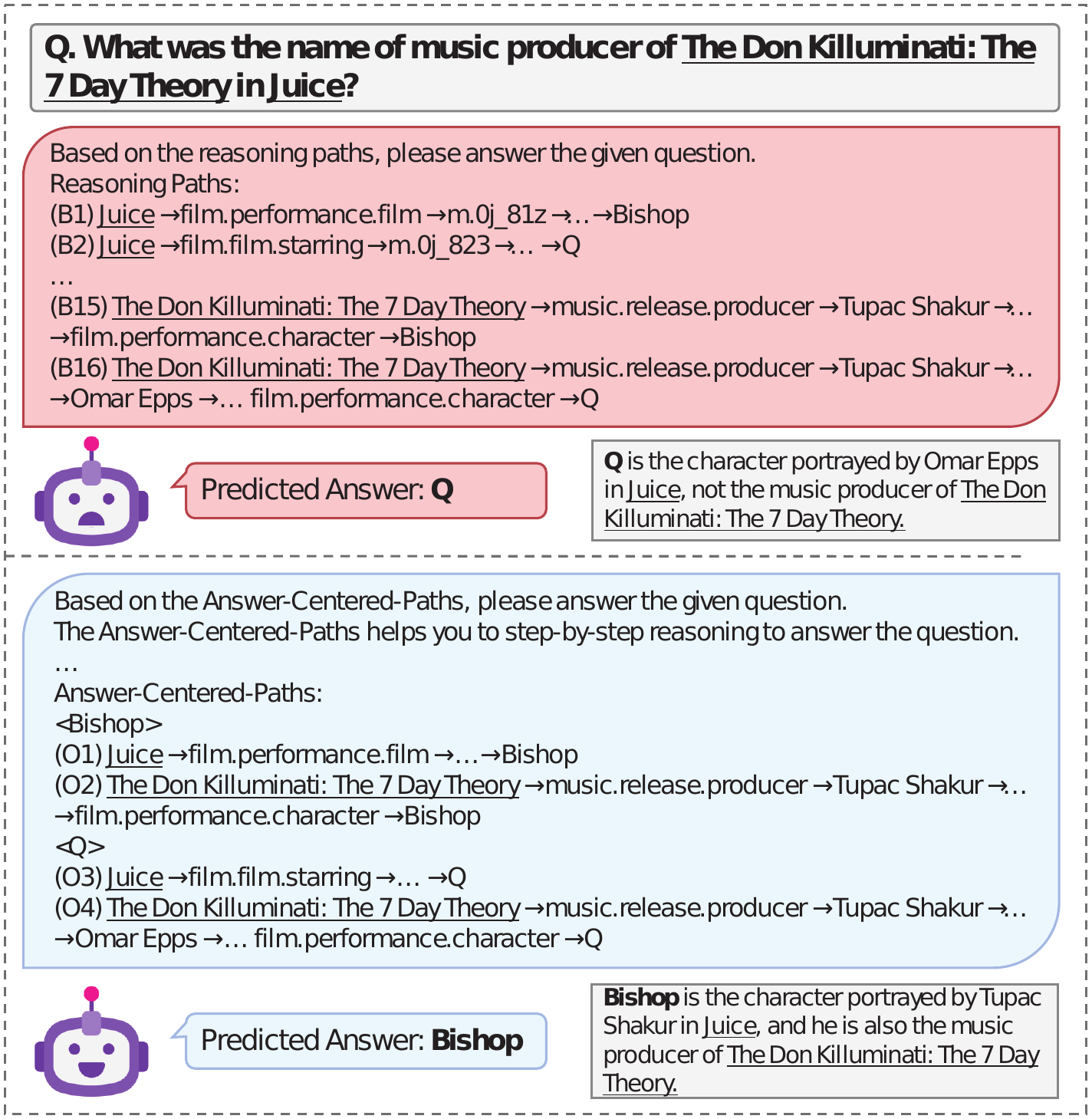}
  \caption{An example of the prompt used in existing works (top) and our designed prompt (bottom), each shown with a predicted answer from a small LLM (Llama3.2-3B \protect\footnotemark[1]). Paths (B1)–(B16) correspond to reasoning paths in existing works, while (O1)–(O4) denote our answer-centered reasoning paths. \textbf{Angle brackets mark candidate answers} (e.g., \texttt{<Bishop>} and \texttt{<Q>}); (O2) and (O3) group paths under these candidates, respectively.}
  \label{fig:motivating_example} 
  \Description{Motivating Example}
\end{figure}
\footnotetext[1]{\url{https://huggingface.co/meta-llama/Llama-3.2-3B-Instruct}}

\section{Introduction}
Large language models (LLMs) have achieved impressive performance across a wide range of NLP tasks ~\cite{kojima2022large, qiao2023reasoning, wei2022chain} but remain vulnerable to hallucination ~\cite{huang2023large, ji2023survey} in knowledge-intensive tasks. Retrieval-augmented generation (RAG) mitigates this issue by grounding generation in external knowledge sources (e.g., documents, databases) ~\cite{shuster2021retrieval, wu2023retrieve}. Among them, knowledge graphs (KGs) are particularly appealing due to their structured representation of factual and relational knowledge, providing a natural foundation for complex reasoning tasks such as question answering ~\cite{ding2024enhancing, xu2025harnessing, liu2025symagent} and conversational recommender system ~\cite{wang2023improving, friedman2023leveraging}. However, when RAG leverages KGs—so-called KG-based RAG~\cite{edge2024local, pan2024unifying}—the key challenge lies in bridging symbolic graph reasoning with the text-based reasoning capabilities of LLMs.

Existing KG-based RAG approaches can be broadly divided into two lines. A first group ~\cite{luoreasoning, luograph} leverages LLMs themselves to plan or retrieve knowledge from KGs by generating candidate path sequences given a query, which are then grounded in the KG before reasoning. A second group ~\cite{lisimple, mavromatis2025gnn} adopts lightweight retrievers such as graph neural networks (GNNs) to extract relevant knowledge. These methods are more efficient than LLM–based retrieval and naturally exploit graph structure, making them less vulnerable to hallucination and suitable for complex multi-hop reasoning.

However, these approaches still face two fundamental challenges. First, their path sampling method for training data construction is typically semantics-unaware, relying on shortest-path heuristics (e.g., BFS) that often select paths inconsistent with a query’s intent. These irrelevant paths are then used as supervision signals, causing models to internalize misaligned reasoning patterns. Consequently, models tend to prioritize topological proximity along paths rather than semantic relevance to the query.

Second, there exists a weak alignment between the retrieved paths and the reasoning objectives of LLMs. While large-scale models (e.g., GPT-4) can partially offset retrieval noise using their extensive parametric knowledge, small LLMs (1–8B parameters) are far more sensitive to both retrieval noise and ungrouped path evidence due to their limited reasoning capacity. As retrieved paths are presented as flat lists rather than answer-centered reasoning paths, small models receive little guidance for integrating them into coherent reasoning. Moreover, the training objective focuses solely on predicting the final answer from consecutive paths, which neglects explicit supervision of intermediate step-by-step reasoning. This mismatch weakens supervision and leads to unstable reasoning performance. 

Figure~\ref{fig:motivating_example} illustrates how unordered retrieval supervision distorts reasoning behavior—small LLMs fail to compose semantically relevant paths even when the correct evidence exists in the KG. The query asks for the character in \textit{Juice} who is also the music producer of \textit{The Don Killuminati: The 7 Day Theory} and the answer is \textit{Bishop}. To satisfy both constraints, the reasoning process must identify the person who serves as the bridge between the two domains — the album (through the \texttt{producer} relation) and the movie (through the \texttt{performance} relation) — thereby linking both domains into a coherent reasoning path. Prior prompts provide retrieved KG paths as an ungrouped list (e.g., path (B2)), which are topologically close to the topic entity \textit{Juice} but ignore the album constraint. Such unorganized evidence biases the model toward frequent but incorrect entities such as \textit{Q}. In contrast, our approach reconstructs retrieved paths into answer-centered reasoning paths that preserve semantic intent (e.g., path (O2)). This representation enables small LLMs to reason over coherent evidence rather than isolated fragments.

Building on this idea, we propose \textbf{RPO-RAG} (\textbf{R}elation-aware weighted \textbf{P}re\-ference \textbf{O}ptimization for RAG), a framework tailored to small LLMs. Unlike prior works, RPO-RAG refines supervision signals across the entire retrieval–reasoning pipeline to better align with the structure of KGs. First, we introduce a query-path semantic sampling strategy that replaces heuristic path construction with a similarity-based approach. By selecting reasoning paths based on semantic consistency with the query, RPO-RAG builds a high-quality training dataset that provides precise supervision for both retriever and reasoner. Second, RPO-RAG employs a relation-aware weighted preference optimization objective that explicitly supervises intermediate reasoning at the relation level. This allows small LLMs to learn fine-grained relational preferences and align their reasoning process with the query intent. Finally, we design an answer-centered prompt that organizes retrieved entities and supporting paths into coherent reasoning contexts, enabling small LLMs to focus on relevant evidence more effectively.
\\
The main contributions of the work are as follows:
\begin{itemize}[leftmargin=*]
    \item We propose RPO-RAG, a KG-based RAG framework tailored to small LLMs. It introduces a query-path semantic sampling strategy that automatically identifies semantic relevance between queries and reasoning paths, enabling the construction of high-quality training dataset with refined supervision signals.
    \item RPO-RAG explicitly models the reasoning process of LLMs at the relation level. To the best of our knowledge, it is the first framework that incorporates relations into preference optimization for KG-based RAG.    
    \item Experimental results on KGQA benchmarks show that our framework consistently outperforms existing methods and, in particular, substantially narrows the performance gap of small-scale LLMs compared to larger-scale baselines.
\end{itemize}

\begin{figure*}[t]
  \centering
  \includegraphics[width=\textwidth,height=0.38\textheight,keepaspectratio]{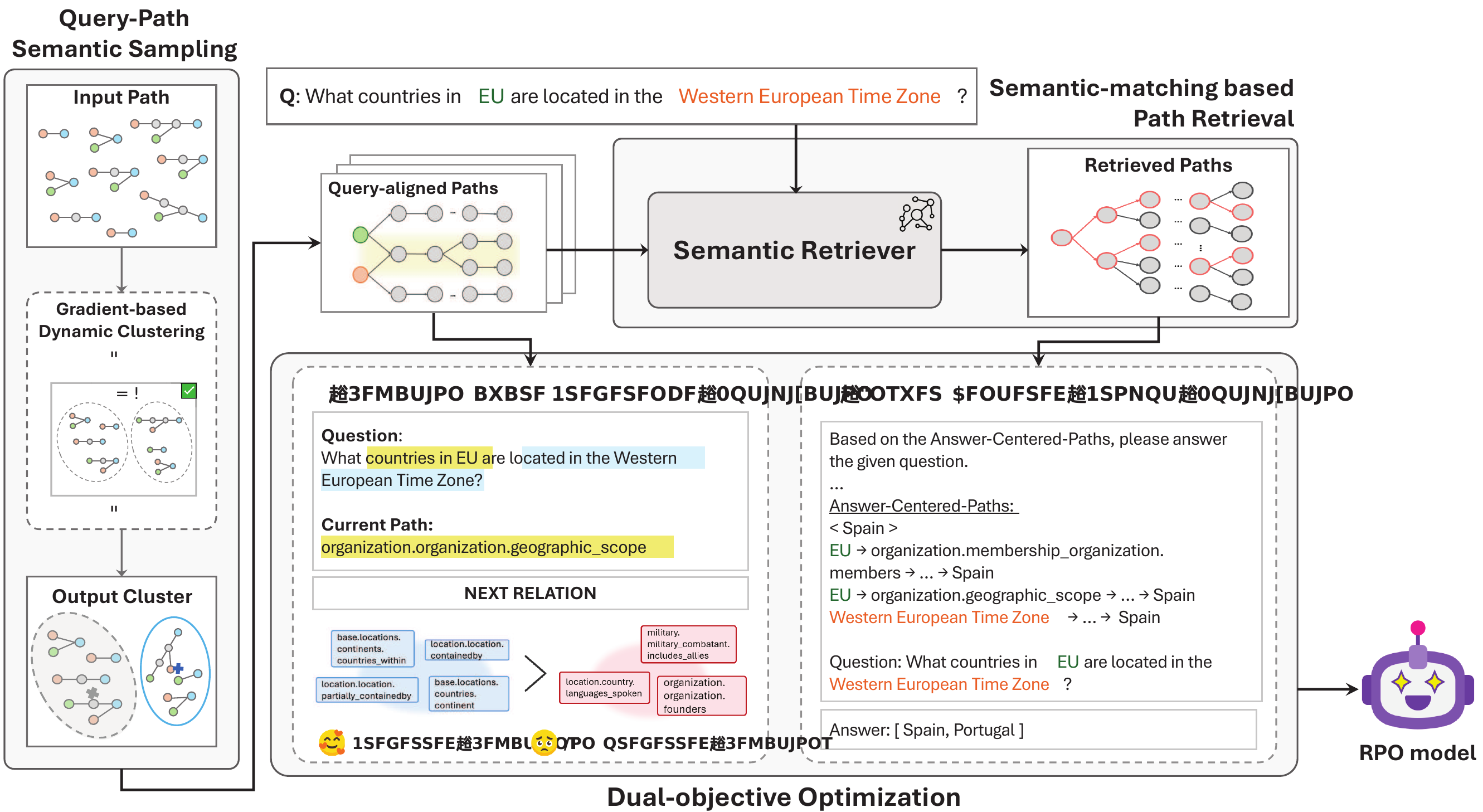}
  \caption{Overview of the RPO-RAG framework. (1) Query-Path Semantic Sampling: constructs query-aligned training paths via dynamic clustering to capture query intent. (2) Semantic-Matching Retriever: retrieves reasoning paths semantically consistent with the query using a pretrained language model. (3) Dual-Objective Optimization: optimizes relation-level preference and answer-centered prompt objectives to align small LLMs with structured reasoning. }
  \Description{Main Architecture}
  \label{fig:teaser}
\end{figure*}
\section{Related Work}
\subsection{Preference Optimization}
Preference optimization (PO) is a training paradigm that optimizes a model’s generation with human preferences by comparing pairs of responses, typically consisting of a preferred $(y^+)$ and a non-preferred $(y^-)$. Early approaches are grounded in Reinforcement Learning from Human Feedback (RLHF)~\cite{christiano2017deep, ouyang2022training}. In this setting, a supervised model is first trained, followed by a separate reward model learned from human-annotated preferences. While effective, this two-stage pipeline suffers from instability and inefficiency, as reinforcement learning is computationally intensive and slow. To improve efficiency, Direct Preference Optimization (DPO) ~\cite{DPO2023} directly optimizes the model such that the probability of generating $y^+$ exceeds that of $y^-$, removing the need for a reward model. However, DPO typically requires a costly reference model to prevent policy collapse. SimPO ~\cite{meng2024simpo} further enhances efficiency by discarding the reference model and employing length correction to mitigate response bias.

Despite these advances, existing PO methods ~\cite{stiennon2020learning, ouyang2022training, DPO2023, ko2024evidence} have been exclusively applied to high-level text generation tasks such as dialogue or summarization, relying heavily on human-annotated data. In contrast, we present an approach that extends preference optimization to relation-level supervision in KGQA. Specifically, we optimize the probability of generating the next relation conditioned on partial paths, aligning the model with the query's intent. Crucially, we construct preference pairs via a weakly supervised method based on semantic relevance of sampled paths, avoiding reliance on costly manual annotations.
\subsection{KG-based RAG}
Integrating KGs into RAG has emerged as a promising strategy to mitigate hallucination in knowledge-intensive tasks. Early works such as SR~\cite{zhang2022subgraph}, NSM\cite{he2021improving}, and UniKGQA~\cite{jiangunikgqa} leverage pretrained language models (PLMs) or graph neural networks (GNNs) to retrieve question-relevant subgraphs without relying on LLMs in the reasoning phase. These methods achieve fast retrieval and reasoning efficiency, but struggles with compositional multi-hop reasoning that requires deeper semantic understanding.

Building on this line, recent KG-based RAG methods can be categorized by whether LLMs are directly involved in the retrieval stage. Some works ~\cite{luoreasoning, luograph} leverage the generative capacity of LLMs to plan or retrieve knowledge from KGs. RoG ~\cite{luoreasoning} employs a planning-retrieval-reasoning pipeline where the LLM proposes candidate relation sequences later grounded in the KG. Similarly, GCR ~\cite{luograph} introduces KG-Trie as a constraint to enforce relation ordering during path extraction, thereby reducing hallucinations. While these methods leverage the reasoning ability of LLMs, they also incur substantial inference overhead due to the complexity of LLM-based path generation. 

To improve efficiency, other approaches avoid LLM inference during retrieval. SubgraphRAG ~\cite{lisimple} adopts an MLP-based parallel triple extraction scheme with distance-based encoding, effectively capturing entities near the topic entity but struggling with multi-hop reasoning. GNN-RAG ~\cite{mavromatis2025gnn} instead frames KGQA as a node classification task, first predicting candidate answers via GNN propagation and then extracting paths to those predicted candidates. Despite these advances, both lines of work share notable limitations: reliance on shortest-path heuristics that ignore query-path semantics, and limited modeling of intermediate relational reasoning caused by flat, ungrouped prompt design, which is particularly detrimental for smaller LLMs. 

Within this landscape, our framework aligns with efficiency-oriented approaches by employing a PLM-based semantic retriever. Its novelty lies in introducing adaptive, semantics-aware path sampling and relation-level optimization.
\section{Method}
\textbf{RPO} (\textbf{R}elation-aware \textbf{P}reference \textbf{O}ptimization)\textbf{-RAG} is a novel framework designed to enhance the reasoning capabilities of small LLMs in the KG-based RAG paradigm. The framework adopts an end-to-end retrieval-and-reasoning pipeline that refines learning signals across the system. It consists of three main components. \textbf{(1) Query-Path Semantic Sampling} constructs a high-fidelity dataset that captures query intent and provides supervisory signals for both retriever and reasoner. \textbf{(2) Semantic-Matching Retriever} trained on this sampled dataset employs dynamic beam search to efficiently extract semantically relevant reasoning paths. Finally, \textbf{(3) Dual-Objective Optimization} integrates \textit{relation-aware relevance-weighted preference optimization} with \textit{answer-centered prompt optimization}, substantially improving the reasoning ability of small LLMs. The overall architecture is illustrated in Figure ~\ref{fig:teaser}.
\subsection{Query-Path Semantic Sampling}
\label{qp-sampling}
\begin{figure}[t]
  \centering
  \includegraphics[width=\columnwidth,keepaspectratio]{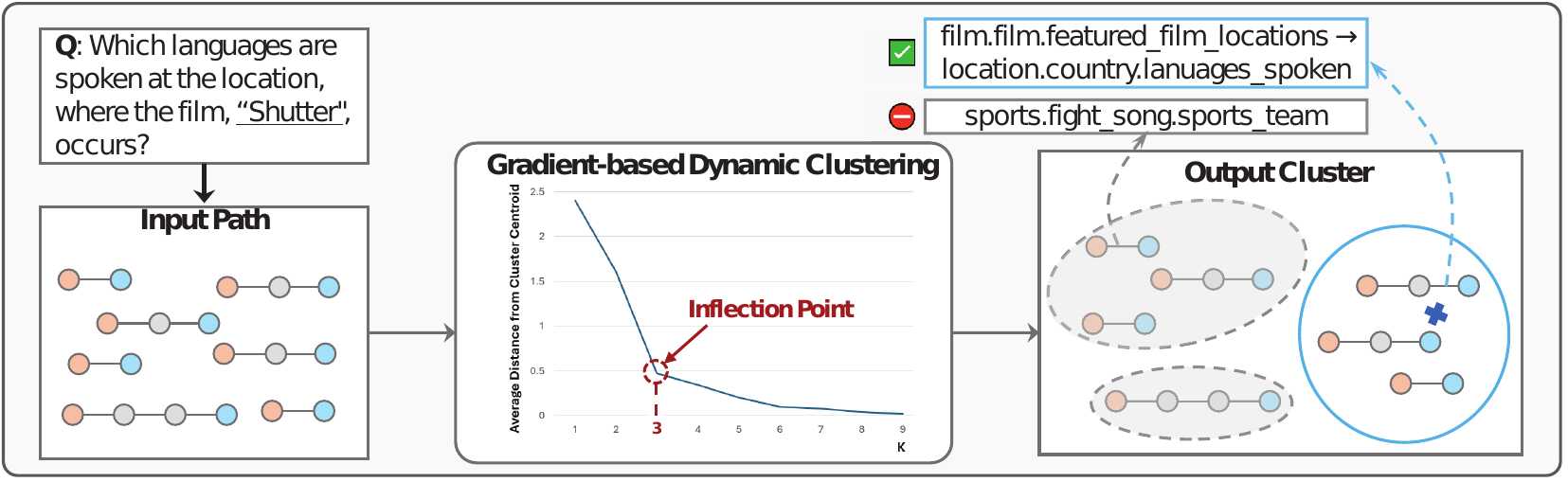}
  \caption{An example of Query-Path Semantic Sampling.}
  \Description{Overview of Query-Path Semantic Sampling.}
  \label{fig:query-path}
\end{figure}

The purpose of query-path semantic sampling is to construct training data that captures query intent, providing reliable supervision for both retriever and reasoner. Prior studies that rely on shortest-path heuristics (e.g., BFS) often include semantically irrelevant paths and fail to reflect the varying reasoning semantics across queries. In contrast, our approach dynamically identifies query-aligned paths by leveraging a gradient-based dynamic clustering, as illustrated in Figure ~\ref{fig:query-path}. This process allows the model to automatically adjust to varying query complexities, rather than depending on pre-defined sampling rules.\\
\textbf{Formalization and Procedure} \\
Given a query $q$, we denote the topic entity by $e_q$ and the answer entity by $e_a$. The set of candidate paths connecting them is defined as:
\begin{equation}
  P = \{p \mid p = (e_q; r_1; r_2; \dots; e_a)\}
\end{equation}
Candidate paths are initially obtained by enumerating all possible shortest paths between $e_q$ and $e_a$ in KG. Each query and path is embedded with a PLM, producing vectors $h_q=PLM(q) $ and $h_p = \text{PLM}(p)$. The semantic relevance between a query and a path is measured by cosine similarity:
\begin{equation}
    s(q,p) = sim(h_q, h_p)
\end{equation}
To capture varying degrees of relevance, we apply a gradient-based dynamic clustering algorithm ~\cite{satopaa2011finding} that partitions paths according to their similarity distribution. This method adaptively determines the optimal number of clusters $N$ by identifying inflection points where the inter-cluster similarity differences are maximized. Concretely, $N$ is determined at the point of maximum curvature in the similarity–cluster curve, effectively detecting semantic boundaries among reasoning paths. From the resulting clusters, the representative cluster $\mathcal{C}^*$ whose embedding $c_k$ is most similar to the query embedding is selected:
\begin{equation}
    \mathcal{C}^* = \arg\max_{C_k} \text{sim}(h_q, c_k)
\end{equation}
The final high-fidelity training set is then defined as:
\begin{equation}
    \hat{P}(q, e_q, e_a) = \{ p \in P \mid p \in \mathcal{C}^* \}
\end{equation}
\textbf{Example}\\
Figure~\ref{fig:query-path} illustrates the procedure with an example. 
Given the query \textit{“Which languages are spoken at the location where the film ‘Shutter’ occurs?”}, 
the topic entity is \textit{Shutter} and the answer entity is \textit{Thai language}. 
Initially, all shortest paths between these entities are enumerated, including both semantically relevant and irrelevant ones, such as:
\begin{itemize}[leftmargin=*]
\item \texttt{film.film.\allowbreak featured\_film\_locations \allowbreak → \allowbreak location.country.\allowbreak languages\_spoken} (relevant)
\item \texttt{film.film.\allowbreak subjects \allowbreak → \allowbreak sports.fight.\allowbreak song.\allowbreak sports\_team} (irrelevant)
\end{itemize}
The gradient-based clustering then identifies an inflection point at $K=3$, grouping paths by semantic similarity. The selected cluster (rightmost in Figure~\ref{fig:query-path}) retains only those paths that align with the query intent, filtering out irrelevant connections. This process enables the training data to capture the precise reasoning semantics of each query and reflects reasoning complexity, providing cleaner and more consistent supervision for both the retriever and the reasoner. We further verify the quality improvement of the sampled data in Section~\ref{data-quality}.
\subsection{Semantic-Matching based Path Retrieval}
The retriever is designed to identify reasoning paths that align semantically with the query intent and to deliver them effectively to the reasoning module. It is trained on the dataset obtained from query-path semantic sampling (Section ~\ref{qp-sampling}), enabling it to encode contextual information that strengthens query-to-path matching. 
\\
\textbf{Retrieval Scheme}
\\
Following SR~\cite{zhang2022subgraph}, we pretrain a PLM using weak supervision from $(q, a)$ pairs. The model learns to predict correct relations along query-answer paths while minimizing sampled negatives, thereby capturing semantic alignment between questions and reasoning paths.

At inference time, the retriever computes semantic similarities between the query and the candidate paths based on aggregated scores. The probability of expanding a relation $r_i$ given the current query $q$ and partial path $p_{t-1}$ is formalized as~\cite{zhang2022subgraph}: 
\begin{equation}
    p(r | q^{(t)}) = \frac{1}{1 + \exp (s(q^{(t)}, END) - s(q^{(t)}, r))},
\end{equation}
where \texttt{END} denotes a virtual relation representing the termination of path expansion. \\
\textbf{Efficient and Noise-Resistant Path Filtering}
\\
Instead of expanding a fixed number of paths as in SR, we employ a dynamic beam search. The expansion size is dynamically adjusted by a threshold value that reflects the gap between similarity scores, where larger gaps trigger pruning of less relevant candidates. This adaptive design prevents redundant path expansion and ensures retrieval that is both efficient and semantically faithful.

To filter out paths irrelevant to the query intent, we constrain the retrieval phase using entity type information from the KG. Structured KGs like Freebase \cite{Freebase} and DBpedia \cite{DBpedia} provide rich type schemas, and answer entity type prediction helps to identify paths that do not meet the query intent.
Firstly, for each entity \(e\in \mathcal{E}\), we retrieve a set of entire entity types \(T_u=\left\{ \tau_1, \tau_2, \ldots,\tau_{|T_u|}\right\}\) from the schema. Afterwards, retrieved entity types are labeled as 1 if connected with answer entities, otherwise 0. We train our model with those labels to predict the answer entity types. 
Specifically, we define a relevance score between each question \(q\) and type \(\tau\) as:
\begin{equation}
\hat{m}_{q,\tau}=ReLU(\textbf{W}_q\textbf{h}_q \times \textbf{W}_{\tau}\textbf{h}_{\tau}),
\end{equation}
where \(\textbf{W}_q, \textbf{W}_\tau\) are projection matrices, and \(\textbf{h}_q, \textbf{h}_\tau\) are embeddings of \(q\) and \(\tau\), respectively.
Finally, we optimize the following cross-entropy loss to predict top-\(K\) answer entity types for each question:
\begin{equation}
\begin{split}
\mathcal{L}_{Type}=-\sum_{q\in Q}\sum_{\tau \in T_q}(m_{q, \tau}\log(\hat{m}_{q,\tau})+\\
(1-m_{q, \tau})\log(1-\hat{m}_{q,\tau})),
\end{split}
\end{equation}
where \(m_{q, \tau}\) is a labeled entity type score. After training, we use the top-5 prediction result to exclude paths whose terminal entity is inconsistent with the predicted types.

\subsection{Dual-Objective Optimization}
\begin{figure}[h]
  \centering
  \includegraphics[width=1\linewidth, height=10cm, keepaspectratio]{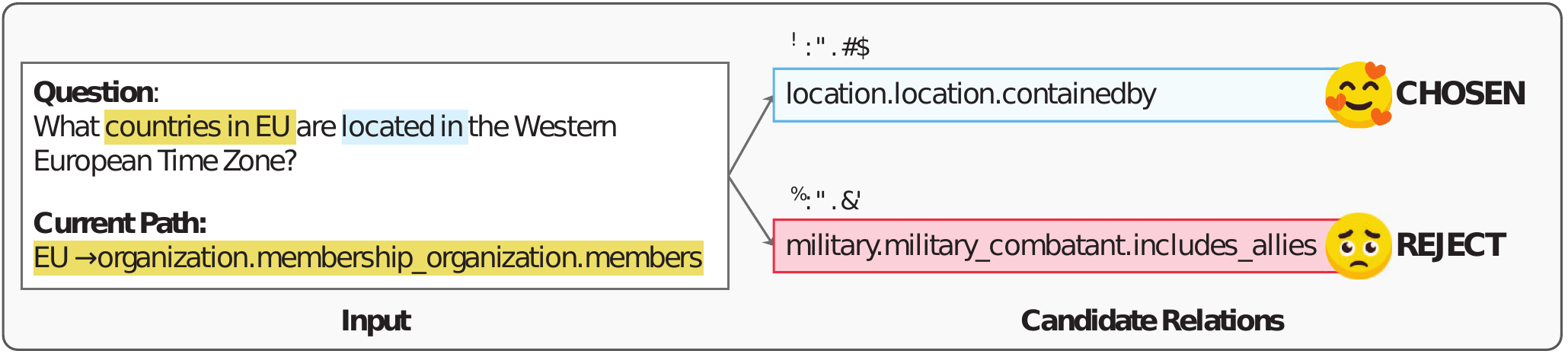}
  \caption{Illustration of Relation-aware Weighted Preference Optimization. 
  For the same question and current path (left), candidate next relations (right) are scored by semantic relevance. Higher-scored relations are treated as preferred (\texttt{CHOSEN}), while semantically misaligned ones are treated as non-preferred (\texttt{REJECT}).}
  \label{fig:relation_aware}
  \Description{Relation-aware Weighted Preference Optimization}
\end{figure}

\textbf{Relation-aware Weighted Preference Optimization} 
\\
A key novelty of our framework is, to our knowledge, the first application of preference optimization at the relation level for knowledge graph reasoning. While prior works mainly focus on optimizing answers or paths as whole units, we introduce a fine-grained objective that supervises the inference of intermediate relations. This design explicitly guides small LLMs to reason over structured relation sequences step by step, rather than only focusing on end entities. As illustrated in Figure~\ref{fig:relation_aware}, our model learns to prefer relations semantically consistent with the query context. As shown in the example, these relation-level preference signals encourage step-by-step, semantically grounded reasoning.

Since the importance of each relation varies across queries, we assign adaptive confidence weights based on semantic relevance. To construct preference signals, we first identify positive and negative relation sets. Relations from the representative cluster $\mathcal{C}^*$ obtained during path sampling are regarded as preferred responses $Y^+$, while relations from alternative clusters are treated as non-preferred responses $Y^-$. This weakly supervised construction enables preference pairs without annotation.

Each relation is weighted according to its semantic proximity to the cluster centroid. Preferred relations closer to the centroid receive higher confidence, while non-preferred relations farther away are penalized. Formally, letting $u(y)$ denote the centroid distance and $\alpha>0$ a decay rate, we define:
\begin{equation}
s(y) =
\begin{cases}
e^{-\alpha\, u(y)} & \text{if } y \in Y^+ \;\; \text{(preferred)}\\[2pt]
1 - e^{-\alpha\, u(y)} & \text{if } y \in Y^- \;\; \text{(non-preferred)}
\end{cases}
\end{equation}
These confidence scores are transformed into weights $(w^+, w^-)$ with a scaling factor $\beta>0$:
\begin{equation}
w = \beta \big(1 + 0.5(s - 0.5)\big)
\end{equation}
Training then proceeds with a margin-based preference objective:
\begin{equation}
\mathcal{L}(\pi_\theta) = - \mathbb{E}\Big[\log \sigma\big( 
W^+ \log \pi_\theta(y^+ \mid x) - 
W^- \log \pi_\theta(y^- \mid x) - \gamma 
\big)\Big],
\end{equation}
where $W^+ = w^+ / |y^+|$ and $W^- = w^- / |y^-| $ are normalized by relation length, and $\gamma$ is a margin term.
By explicitly applying relation-level preference optimization with relevance-weighted signals, our approach introduces a new training paradigm for KGQA. This enables small LLMs to internalize structured, step-by-step relational reasoning and improve their capacity to execute faithful, interpretable inference over KGs.\\
\textbf{Answer-Centered Prompt Optimization} \\
The second objective complements relation-level training by aligning answer generation with answer-centered reasoning paths. \allowbreak Prompts are explicitly designed to incorporate multiple reasoning paths as evidence supporting candidate answers. Unlike conventional prompts that treat each path in isolation and fail to integrate information from multiple topic entities, our proposed prompt unifies these paths into a coherent representation better suited for small LLMs. As illustrated in Figure~\ref{fig:motivating_example}, we adopt an \textit{answer-centered} layout: paths are grouped by their end entities, and all paths supporting the same candidate are presented together. For multi-topic questions, this structure naturally merges reasoning paths originating from different topics but converging on the same answer, allowing small LLMs to aggregate consistent evidence more effectively.  

Formally, we maximize the likelihood of the correct answer conditioned on the answer-centered prompt: 
\begin{equation}
\mathcal{L}_{\text{ans}} = - \mathbb{E}_{(q,a)} \, \log P_{\theta}(e_a \mid x_{\text{answer-centered}})
\end{equation}
where $x_{\text{answer-centered}}$ denotes the grouped (answer-centered) prompt. This optimization not only improves answer accuracy but also enhances interpretability, as predictions are explicitly tied to transparent reasoning paths.

\begin{table*}[t]
  \caption{
    Main results on WebQSP and CWQ. 
    \textbf{RPO-RAG} achieves strong improvements among small LLMs ($\leq$8B)
    and significantly narrows the gap with GPT-based models. \small
  $^{\dagger}$Reproduced using the official implementation released by the original authors.
  }
  \label{tab:main_results}
  \centering
  \setlength{\tabcolsep}{12pt}
  \renewcommand{\arraystretch}{1.05}
  \begin{tabular}{l l cc cc}
    \toprule
    \multirow{2}{*}{\textbf{Type}} & \multirow{2}{*}{\textbf{Methods}} &
    \multicolumn{2}{c}{\textbf{WebQSP}} & \multicolumn{2}{c}{\textbf{CWQ}} \\
    \cmidrule(lr){3-4}\cmidrule(lr){5-6}
     & & \textbf{Hit} & \textbf{F1} & \textbf{Hit} & \textbf{F1} \\
    \midrule

    \multirow{4}{*}{GNN-based}
      & GraftNet              & 66.7 & 62.4 & 36.8 & 32.7 \\
      & NSM                   & 68.7 & 62.8 & 47.6 & 42.4 \\
      & SR + NSM              & 68.9 & 64.1 & 50.2 & 47.1 \\
      & UniKGQA               & 77.2 & 72.2 & 51.2 & 49.1 \\
    \midrule

    \multirow{4}{*}{Vanilla LLM}
      & Llama2-7B                   & 56.4 & 36.5 & 28.4 & 21.4 \\
      & Llama3.1-8B                 & 55.5 & 34.8 & 28.1 & 22.4 \\
      & Llama3.2-3B                 & 62.5 & 32.2 & 19.9 & 14.4 \\
      & Llama3.2-1B                 & 50.2 & 28.9 & 14.3 & 11.2 \\
    \midrule

    \multirow[c]{9}{*}{KG + LLM}
      & RoG (Llama2-7B)             & 85.7 & 70.8 & 62.6 & 56.2 \\
      & GNN-RAG (Llama2-7B)         & 85.7 & 71.3 & 66.8 & \uline{59.4} \\
      & GNN-RAG (Llama3.1-8B$^{\dagger}$)         & 85.6 & 70.3 & 66.6 & 58.0 \\
      & SubgraphRAG (Llama3.1-8B)   & 86.6 & 70.6 & 57.0 & 47.2 \\
      & GCR (Llama3.1-8B$^{\dagger}$)           & 87.2 & 69.1 & 60.5 & 50.4 \\
      \cmidrule(lr){2-6}
      & RPO-RAG (Llama2-7B)   & \uline{88.3} & \uline{77.8} & \uline{68.1} & 59.2 \\
      & RPO-RAG (Llama3.1-8B) & \textbf{89.9} & \textbf{81.3} & \textbf{72.3} & \textbf{64.5} \\
      &  RPO-RAG (Llama3.2-3B) & 87.3 & 76.4 & 66.0 & 57.3 \\
      & RPO-RAG (Llama3.2-1B) & 82.3 & 69.8 & 60.3 & 50.4 \\
    \bottomrule
  \end{tabular}
\end{table*}

\section{Experiments}
In this section, we presented experimental results and analyses to evaluate the effectiveness of our proposed RPO-RAG framework. 
We aimed to answer the following research questions:\\
\textbf{RQ1:} How effectively and efficiently does RPO-RAG improve overall KGQA reasoning performance?  \\
\textbf{RQ2:} Can small LLMs in RPO-RAG narrow the reasoning gap with large-scale models (e.g., GPT-4o-mini)?\\
\textbf{RQ3:} How do the individual components contribute to the overall performance? \\
\textbf{RQ4:} Does proposed data sampling method enhances data quality and retriever supervision?  
\subsection{Experimental Settings} 
\textbf{Datasets.} We evaluated RPO-RAG on two widely used KGQA benchmarks, WebQuestionsSP (WebQSP)~\cite{yih2016value} and Complex WebQuestions (CWQ)~\cite{talmor2018web}, both grounded in Freebase~\cite{Freebase}. WebQSP mainly contains relatively simple 1–2 hop questions, whereas CWQ includes more complex ones requiring up to 4-hop reasoning. Detailed dataset statistics are provided in Appendix~\ref{adx:kgqa}.\\
\textbf{Implementation Details.} We utilized Sentence-BERT~\cite{reimers2019sentence} as a default PLM for path retrieval, trained using query–path semantic sampling. During inference, dynamic beam search is applied with adaptive thresholds for WebQSP and CWQ. The reasoner employs small LLMs—Llama2-7B~\cite{touvron2023llama}, Llama3.1-8B~\cite{dubey2024llama}, Llama3.2-3B\footnotemark[1], and Llama3.2-1B\footnotemark[2]—fine-tuned with LoRA on 2 $\times$ NVIDIA RTX 4090 GPUs. More implementation details are available in Appendix~\ref{adx:implement}.\\
\textbf{Baselines}. We compared RPO-RAG against representative models from three categories: 
(1) \textbf{Graph-based reasoning methods} — GraftNet~\cite{sun2018open}, NSM~\cite{he2021improving}, SR+NSM~\cite{zhang2022subgraph}, and UniKGQA~\cite{jiangunikgqa}; 
(2) \textbf{LLM-only methods} — vanilla small and medium LLMs, including 
Llama\-2–7B~\cite{touvron2023llama}, Llama\-3.1–8B ~\cite{dubey2024llama}, Llama3.2-3B\protect\footnotemark[1], and Llama3.2-1B\protect\footnotemark[2]; and
(3) \textbf{LLM+KG methods} — ToG~\cite{sunthink}, RoG~\cite{luoreasoning}, GCR~\cite{luograph}, SubgraphRAG~\cite{lisimple}, and GNN-RAG~\cite{mavromatis2025gnn}. 
Detailed descriptions for all baselines are provided in Appendix ~\ref{append-baseline}.\\
\textbf{Evaluation Metrics.} We adopted Hit and F1 as evaluation metrics. Hit measures whether at least one correct answer appears in the predicted set, while F1 balances precision and recall to provide a comprehensive assessment of answer quality.
\footnotetext[1]{\url{https://huggingface.co/meta-llama/Llama-3.2-3B-Instruct}}
\footnotetext[2]{\url{https://huggingface.co/meta-llama/Llama-3.2-1B-Instruct}}

\subsection{Main Results (RQ1)}
\textbf{Effectiveness} \\
Table~\ref{tab:main_results} presents the overall performance comparison on WebQSP and CWQ datasets. Our proposed RPO-RAG consistently outperformed all small-scale and graph-based baselines, demonstrating the effectiveness of relation-aware and answer-centered prompt optimization.

On WebQSP, RPO-RAG (Llama3.1–8B) achieves 89.9 Hit and 81.3 F1, establishing a new state-of-the-art (SOTA) among all models up to 8B parameters—surpassing the previous best (GCR) by $+$2.7\% Hit and $+$10.2\% F1. RPO-RAG (Llama2–7B) also delivers the second-best overall result with 88.3 Hit and 77.8 F1, confirming the framework’s robustness across architectures. Notably, even the small-scaled RPO-RAG (Llama3.2–3B) exceeds the performance of several larger baselines, demonstrating that RPO-RAG effectively transfers structured reasoning capability to small LLMs.

On the more challenging CWQ dataset, which requires multi-hop reasoning paths, RPO-RAG achieved consistent gains across both Hit and F1. Specifically, RPO-RAG (Llama2–7B) improved Hit $+$1.3\% and F1 $+$4.9\% over GNN-RAG (Llama2–7B), while RPO-RAG (Llama3.1-8B) reached the highest Hit and F1 among all $\leq$8B baselines. These gains highlight that RPO-RAG not only identifies more correct answers but also sustains better coverage and compositional reasoning across hops.

Beyond these strong relative improvements, RPO-RAG also delivers substantial absolute gains over vanilla LLMs without any external reasoning mechanism. For instance, RPO-RAG (Llama3.2–3B) improves Hit by $+$24.8\% on WebQSP and $+$46.1\% on CWQ compared to the base Llama3.2–3B, while RPO-RAG (Llama3.2–1B) yields $+$32.1\% and $+$46\% Hit gains over its vanilla counterpart. These results confirm that our framework effectively injects structured reasoning capability into small LLMs, even when model capacity is highly constrained. Overall, RPO-RAG substantially enhances the reasoning reliability and scalability of small open-weight LLMs across both benchmarks.\\
\textbf{Efficiency}\\
To assess the efficiency of RPO-RAG, we measured average per-query retrieval and reasoning times on the CWQ, a benchmark that requires complex multi-hop reasoning. Results in Table ~\ref{tab:efficiency_cwq} were obtained on a single NVIDIA RTX 3090 GPU. 

For the retrieval stage, existing methods adopt varying strategies. GNN-RAG and SubgraphRAG use lightweight neural retrievers—graph propagation with a GNN and an MLP-based embedding similarity scorer (SBERT embeddings), respectively. GCR, in contrast, relies on an LLM to explicitly generate candidate reasoning paths. RPO-RAG employs a compact PLM trained on semantically sampled data, which efficiently encodes query–path relevance. On the reasoning side, GNN-RAG and RPO-RAG use fine-tuned LLMs, whereas SubgraphRAG and GCR use vanilla LLMs without task-specific adaptation.

Among the compared methods, RPO-RAG offers the best accuracy–latency balance. SubgraphRAG attains the fastest retrieval but suffers from the slowest end-to-end time and the lowest Hit. GCR shows the opposite pattern: LLM-based retrieval dominates latency. GNN-RAG is the fastest overall but falls behind in accuracy. By contrast, RPO-RAG keeps retrieval lightweight and achieves near-minimal total latency while delivering the highest Hit, providing the strongest accuracy–efficiency trade-off on CWQ.

\subsection{Competitiveness of Small LLMs (RQ2)}
This section evaluates whether the proposed RPO-RAG framework enables small LLMs to approach the reasoning capability of GPT-based models. Table~\ref{tab:rq2_results} summarizes the comparison results across both datasets. Overall, RPO-RAG substantially boosts the reasoning performance of small LLMs, closing the gap with large proprietary LLMs while maintaining a fraction of their parameter scale. 

On the relatively simple WebQSP dataset, which mostly involves up to 2-hop reasoning, even the smallest variant, RPO-RAG (Llama3.2\allowbreak{}–1B), achieves a Hit score of 82.3, surpassing ToG (ChatGPT) by ($+$6.1\%). This demonstrates that relation-aware preference optimization allows a 1B-parameter model to reach GPT-level accuracy despite its limited capacity. 

On the more challenging CWQ dataset, which requires up to 4-hop reasoning, RPO-RAG (Llama3.1–8B) achieves 72.3 Hit and 64.5 F1, approaching the performance of GCR (ChatGPT) and reducing the gap to GCR (GPT-4o-mini) to within roughly 3–4 points. These results collectively confirm that RPO-RAG successfully transfers reasoning consistency and compositional understanding to smaller LLMs. By aligning intermediate reasoning through relation-level preference optimization, our framework enables small, open-weight models to achieve GPT-comparable reasoning accuracy under realistic computational constraints.
\begin{table}[t]
  \centering
  \caption{Efficiency and Hit comparison of different methods on CWQ. Average per-question wall-clock time (seconds). Hit ($\uparrow$) is higher-is-better; Retrieval, Reasoning, and Total ($\downarrow$) are lower-is-better.}
  \label{tab:efficiency_cwq}
  \footnotesize
  \resizebox{\linewidth}{!}{
  \renewcommand{\arraystretch}{1.15}
  \begin{tabular}{@{}lcccc@{}}
    \toprule
    \textbf{Methods} & \textbf{Hit} & \textbf{Retrieval} & \textbf{Reasoning} & \textbf{Total} \\
    \midrule
    GNN-RAG (Llama2-7B)       & 66.8 & 0.11 & 0.95 & \textbf{1.06} \\
    SubgraphRAG (Llama3.1-8B) & 57.0 & 0.02 & 6.12 & 6.14 \\
    GCR (Llama3.1-8B)         & 60.5 & 6.84 & 0.56 & 7.4 \\
    \midrule
    \textbf{RPO-RAG (Llama2-7B)}   & \uline{68.1} & \multirow{2}{*}{0.07} & 1.36 & 1.43 \\
    \textbf{RPO-RAG (Llama3.1-8B)} & \textbf{72.3} &                       & 1.1  & \uline{1.17} \\
    \bottomrule
  \end{tabular}}
\end{table}

\begin{table}[h]
  \caption{
    Comparison of RPO-RAG with GPT-based models on WebQSP and CWQ.
  }
  \label{tab:rq2_results}
  \centering
  \setlength{\tabcolsep}{4pt}
  \renewcommand{\arraystretch}{1.05}
  \begin{tabular}{l cc cc}
    \toprule
    \multirow{2}{*}{\textbf{Methods}} &
    \multicolumn{2}{c}{\textbf{WebQSP}} &
    \multicolumn{2}{c}{\textbf{CWQ}} \\
    \cmidrule(lr){2-3}\cmidrule(lr){4-5}
     & \textbf{Hit} & \textbf{F1} & \textbf{Hit} & \textbf{F1} \\
    \midrule
    ToG (ChatGPT)              & 76.2 & –    & 57.6 & –    \\
    ToG (GPT-4)                & 82.6 & –    & 68.5 & –    \\
    GCR (ChatGPT)              & 92.6 & 73.2 & 72.7 & 60.9 \\
    GCR (GPT-4o-mini)          & 92.2 & 74.1 & 75.8 & 61.7 \\
    SubgraphRAG (GPT-4o-mini)  & 90.1 & 77.5 & 62.0 & 54.1 \\
    \midrule
    \textbf{RPO-RAG (Llama2–7B)}   & 88.3 & 77.8 & 68.1 & 59.2 \\
    \textbf{RPO-RAG (Llama3.1–8B)} & 89.9 & 81.3 & 72.3 & 64.5 \\
    \textbf{RPO-RAG (Llama3.2–3B)} & 87.3 & 76.4 & 66.0 & 57.3 \\
    \textbf{RPO-RAG (Llama3.2–1B)} & 82.3 & 69.8 & 60.3 & 50.4 \\
    \bottomrule
  \end{tabular}
\end{table}

\subsection{Ablation Study (RQ3)}
Table ~\ref{tab:ablation} shows the ablation results of RPO-RAG. \textit{w/o Relation-aware Optimization} removes the relation-level preference optimization, training the model only with answer-centered prompt. \textit{w/o Answer-Centered Prompt} omits the answer-centered prompt design, instead providing flat lists of retrieved paths. 

The results clearly show that both components are essential to the framework’s overall performance. Removing either component consistently degrades both Hit and F1 across datasets and model sizes. Relation-aware optimization contributes more significantly on WebQSP, suggesting its importance for precise reasoning alignment, while the answer-centered prompt yields larger gains on CWQ, reflecting its effectiveness for multi-hop compositional reasoning.

\begin{table}[t]
  \centering
  \caption{Ablation study on WebQSP and CWQ. Removing either component leads to consistent drops in both Hit and F1.}
  \label{tab:ablation}
  \resizebox{\linewidth}{!}{
  \begin{tabular}{lcccc}
    \toprule
    \multirow{2}{*}{\textbf{Method}} & 
    \multicolumn{2}{c}{\textbf{WebQSP}} & 
    \multicolumn{2}{c}{\textbf{CWQ}} \\
    \cmidrule(lr){2-3}\cmidrule(lr){4-5}
     & \textbf{Hit} & \textbf{F1} & \textbf{Hit} & \textbf{F1} \\
    \midrule
    \textbf{RPO-RAG (Llama2-7B)} & \textbf{88.3} & \textbf{77.8} & \textbf{68.1} & \textbf{59.2} \\
    \hspace{2mm}w/o Relation-aware Optimization & 81.6 & 66.5 & 61.1 & 54.2 \\
    \hspace{2mm}w/o Answer-Centered Prompt          & 80.3 & 65.7 & 58.6 & 47.5 \\
    \midrule
    \textbf{RPO-RAG (Llama3.2-3B)} & \textbf{87.3} &\textbf{ 76.4} & \textbf{66.0} & \textbf{57.3 }\\
    \hspace{2mm}w/o Relation-aware Optimization & 78.8 & 63.4 & 58.1 & 55.9 \\
    \hspace{2mm}w/o Answer-Centered Prompt          & 78.1 & 62.7 & 57.3 & 46.6 \\
    \bottomrule
  \end{tabular}}
\end{table}

\subsection{Dataset Quality Analysis (RQ4)}
\label{data-quality}
\begin{figure}[h]
  \centering
  \includegraphics[width=0.85\linewidth, height=3.5cm, keepaspectratio]{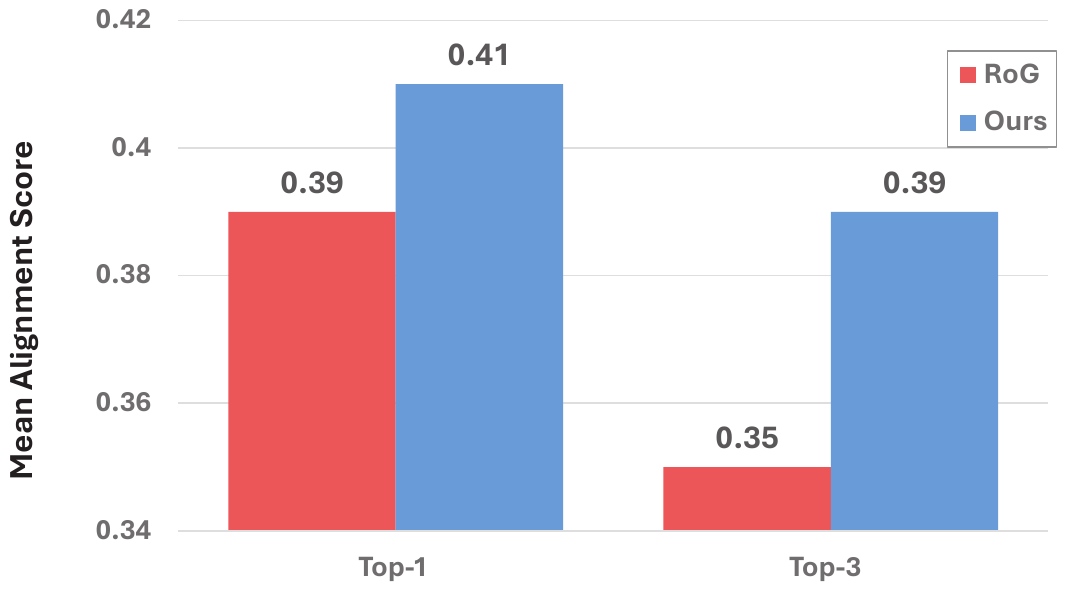}
  \caption{Comparison of query–path semantic alignment between datasets under top-1 and top-3 settings.}
  \label{fig:bar}
  \Description{Comparison of query-path semantic alignment}
\end{figure}

In this section, we evaluated the quality of the dataset constructed using our semantic query–path sampling strategy (Section ~\ref{qp-sampling}).  To assess its effectiveness, we compared against the dataset released by RoG ~\cite{luoreasoning}, which has been widely adopted in prior studies ~\cite{luoreasoning, mavromatis2025gnn, luograph}. Our analysis covered three dimensions: relation coverage, semantic alignment, and improvements in retriever performance. \\
\textbf{Relation Coverage Analysis} \\
We defined ground-truth relations from the annotated SPARQL queries and evaluate how well each dataset reflects the intended semantics of the questions. Specifically, we computed precision, recall, and F1 by comparing the overlap between the ground-truth relation set and the relations contained in the training data. As shown in Table ~\ref{tab:coverage}, our dataset maintains comparable recall while improving precision by $+$7.7\%, leading to a $+$4.3\% increase in F1. These results demonstrate that our sampling strategy effectively filters out noisy paths and preserves semantically aligned relations, thereby providing higher-quality supervision for training. \\
\textbf{Semantic Alignment Analysis} \\
To evaluate semantic consistency, we computed cosine similarity between query and candidate path embeddings using SBERT. For each query, we reported average similarity scores under the \textit{\textit{top-1}} and \textit{top-3} settings. As shown in Figure ~\ref{fig:bar}, while \textit{top-1} results are comparable between datasets, the gap widens to 0.04 at \textit{top-3}. Moreover, the similarity drop from \textit{top-1} to \textit{top-3} is smaller in our dataset (0.03 vs. 0.05 for the baseline). This demonstrates that our strategy maintains stronger semantic alignment with questions even as candidate paths are expanded, highlighting its robustness. \\

\begin{table}[t]
  \caption{Comparison of relation coverage between RoG-cwq and our dataset.}
  \label{tab:coverage}
  \setlength{\tabcolsep}{10pt}
  \centering
  \begin{tabular}{l c c c}
    \toprule
    \textbf{Dataset} & \textbf{Precision} & \textbf{Recall} & \textbf{F1} \\
    \midrule
    RoG-cwq   & 31.5 & 35.2 & 30.8 \\
    Ours-cwq  & \textbf{39.2} & \textbf{35.3} & \textbf{35.1} \\
    \bottomrule
  \end{tabular}
\end{table} 

\noindent{\textbf{Retriever Performance Evaluation}} \\
Building on the previous data-quality verification, we evaluated how the sampled dataset enhances retriever performance by providing stronger supervision signals. As shown in Table ~\ref{tab:retriever_perf}, our retriever achieved a substantial accuracy improvement of \textbf{$+$ 22.4\%} over GNN-RAG, demonstrating that query-aligned supervision leads to more faithful path retrieval. 
Although the average number of retrieved paths (ARP) increases, this reflects broader yet semantically consistent coverage rather than redundancy. In contrast to shortest-path-based retrievers, which tend to miss relevant evidence beyond local neighborhoods, our semantically guided retriever expands reasoning coverage while maintaining precision. Overall, these findings indicate that the proposed dataset enables retrievers to move beyond brittle shortest-path heuristics toward semantically guided retrieval, delivering higher accuracy.

\begin{table}[t]
  \caption{Retriever performance comparison showing accuracy and average retrieved paths (ARP).}
  \label{tab:retriever_perf}
  \centering
  \begin{tabular}{
       @{\hspace{4pt}}l
      @{\hspace{8pt}}c
      @{\hspace{8pt}}c
  }
    \toprule
    \textbf{Model} & \textbf{Accuracy (\%)} & \textbf{ARP}  \\
    \midrule
    RoG       & 69.6    & 27       \\
    GNN-RAG   & 65.0    & 20       \\
    \textbf{Ours} & \textbf{87.4} & \textbf{116}  \\
    \bottomrule
  \end{tabular}
\end{table}

\section{Conclusion}
We presented \textbf{RPO-RAG}, a KG-based RAG framework tailored to small LLMs ($\leq$8B) that (i) samples query–aligned reasoning paths via semantic clustering, (ii) applies relation-aware, relevance-weighted preference optimization to supervise intermediate steps, and (iii) reconstructs prompts to aggregate evidence along answer-centered paths. Across WebQSP and CWQ, RPO-RAG consistently improves Hit and F1 over graph-based and small-LLM baselines, while markedly narrowing the gap to GPT-based systems. Notably, even 1B–3B backbones show substantial gains, indicating that relation-level supervision and answer-centered prompting effectively transfer compositional reasoning to compact models. 

Our dataset-quality analyses further show that semantics-aware sampling yields higher-quality supervision and more faithful retrieval. Taken together, these findings suggest a practical path to scalable, resource-efficient KGQA: align retrieval and reasoning \emph{at the relation level} and deliver evidence in a form that small LLMs can exploit. A more detailed discussion of additional limitations and future
research directions is provided in Appendix~\ref{adx:limitations}.


\bibliographystyle{ACM-Reference-Format}
\bibliography{reference}

\appendix
\section{Limitations and Future Work}
\label{adx:limitations}
While RPO-RAG substantially narrows the gap to larger systems, there remains ample room for improvement on compositional queries that require reasoning over attribute values.
As shown in Section~\ref{adx:error} and Figure~\ref{fig:error_case}, our current pipeline does not explicitly encode or propagate attribute-conditioned evidence, which leads to a higher error rate on such query types.
This limitation is amplified for smaller backbones ($\leq$3B), where the capacity to infer attribute-dependent relations is further constrained.
Future work includes integrating attribute-aware retrieval and path construction (e.g., schema- or property-guided sampling), designing training signals that supervise relation and attribute consistency, and developing prompt or adapter mechanisms that let small LLMs represent and exploit attribute information more reliably.

\section{Experimental Details}
\subsection{KGQA benchmarks}
\label{adx:kgqa}
We utilized two representative KGQA benchmarks: WebQuestionsSP (WebQSP) ~\cite{yih2016value}and ComplexWebQuestions (CWQ) ~\cite{talmor2018web}. 
Both datasets are grounded in the Freebase knowledge graph ~\cite{Freebase}, and their statistics are summarized in Table~\ref{tab:dataset-stats}.
\begin{table}[h]
  \centering
  \caption{Statistics of datasets used in experiments. The number of QA pairs for training and testing is reported along with the maximum reasoning hop.}
  \label{tab:dataset-stats}
  \setlength{\tabcolsep}{10pt} 
  \begin{tabular}{lccc}
    \toprule
    \textbf{Dataset} & \textbf{Train} & \textbf{Test} & \textbf{Max Hop} \\
    \midrule
    WebQSP & 2,848 & 1,628 & 2 \\
    CWQ & 27,639 & 3,531 & 4 \\
    \bottomrule
  \end{tabular}
\end{table}
\subsection{Implementation Details}
\label{adx:implement}
\begin{figure}[h]
    \centering
    \includegraphics[width=\linewidth]{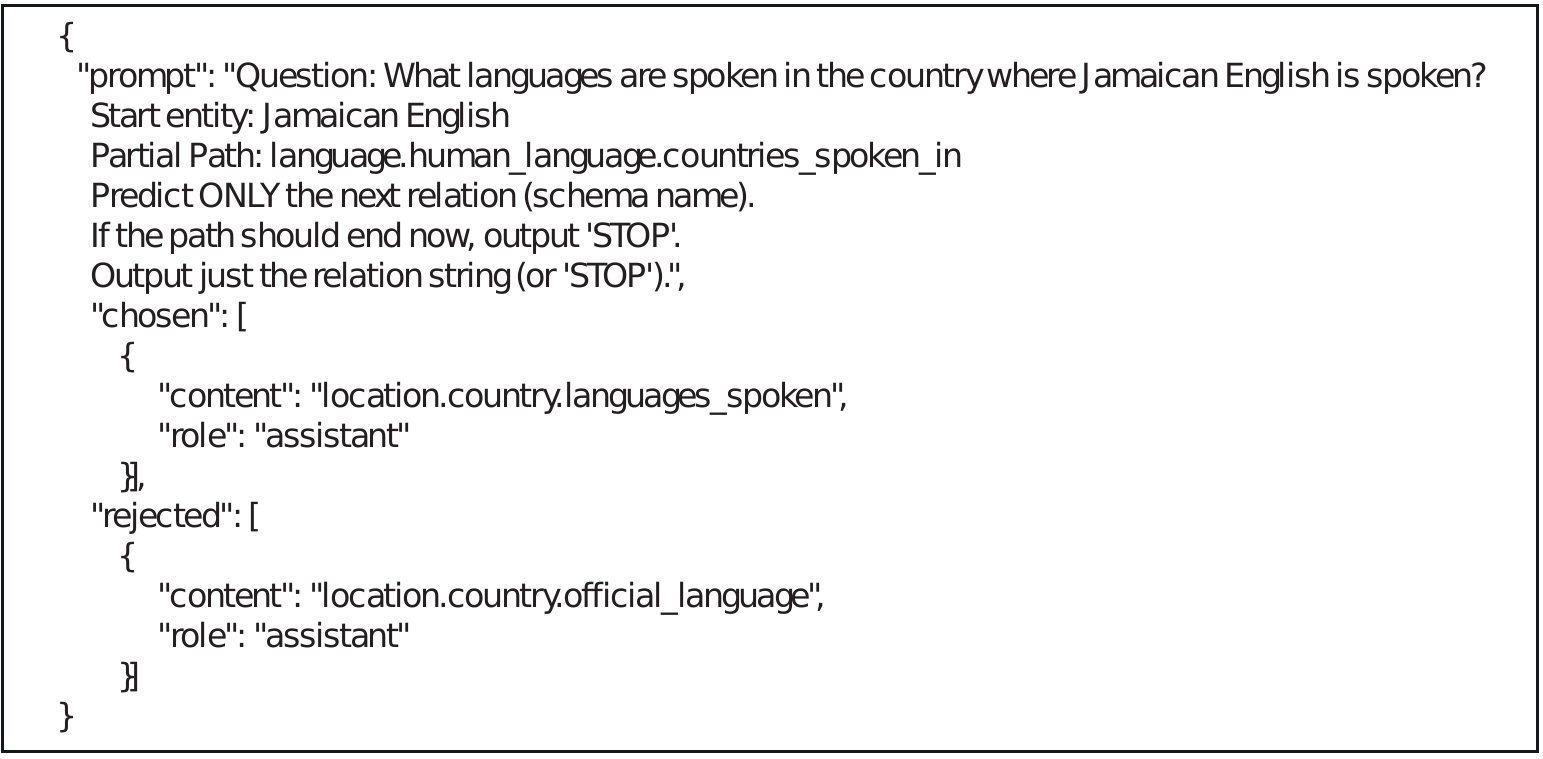}
    \caption{An example train data of Relation-aware Weighted Preference Optimization}
    \label{fig:relation-data}
\end{figure}

\textbf{Query-Path Semantic Sampling}\\
For embedding construction, we use the \textit{all-MiniLM-L6-v2} (22.7M
params) as an encoder. All query and path embeddings are L2-normalized, and cosine similarity is computed via inner product between normalized vectors. To identify semantically coherent path groups, we apply KMeans clustering, switching to MiniBatchKMeans when the number of candidate paths exceeds 1,500. The maximum number of clusters is set to $K_{\max}=10$, and the optimal cluster count $k$ is automatically determined using the elbow criterion, selecting the point of largest inertia drop. All experiments are performed with a fixed random seed of 42 to ensure reproducibility.\\
\textbf{Semantic-Matching based Path Retrieval} \\
We fine-tuned Sentence-BERT as a default retriever. For path expansion, we initialize beam width as 10. For dynamic beam search, we set threshold as 0.3 for both WebQSP and CWQ. For type prediction, we utilized gte-large-en-v1.5 for encoder. We used top-5 type prediction result for path filtering. \\
\textbf{Relation-aware Weighted Preference Optimization} \\
We fine-tune with LoRA on 2 $\times$ NVIDIA RTX 4090 GPUs. The detailed hyperparameters are listed in Table~\ref{tab:detailed_hparams} (dataset-specific LR/epochs for WebQSP vs.\ CWQ). For preference-based optimization, we construct training data that encode question–path pairs, along with preferred (\texttt{chosen}) and non-preferred (\texttt{rejected}) reasoning paths. Each example corresponds to a partial traversal in the knowledge graph, where the model must predict the next relation or terminate the path (\texttt{"STOP"}). An example of such data is illustrated in Figure~\ref{fig:relation-data}.\\
\\
\textbf{Answer-Centered Prompt Optimization} \\
Using the same LoRA configuration, we fine-tune both WebQSP and CWQ for 3 epochs with AdamW-8bit (lr = $2{\times}10^{-4}$), a cosine scheduler with warmup = 0.03. Training is conducted on 2 $\times$ NVIDIA RTX 4090 GPUs. The corresponding hyperparameters are also summarized in Table~\ref{tab:detailed_hparams}.

\subsection{Baselines}
\label{append-baseline}
We evaluated RPO-RAG against representative graph-reasoning and KG-augmented LLM baselines.\\
\textbf{Graph-based Methods}
\begin{itemize}[leftmargin=*]
    \item GraftNet ~\cite{sun2018open} introduces an early graph-neural approach for KGQA that retrieves question-specific subgraphs via entity linking and performs reasoning using a convolution-based GNN.
    \item NSM ~\cite{he2021improving} models multi-hop reasoning as a sequential decision process, predicting relations step-by-step over KGs.
    \item SR+NSM ~\cite{zhang2022subgraph} extends NSM by proposing relation-path retrieval, which selects relation sequences relevant to the question to construct subgraphs for reasoning.
    \item UniKGQA ~\cite{jiangunikgqa} unifies graph retrieval and reasoning into a single framework, employing language modeling to bridge symbolic KG reasoning and text understanding.
\end{itemize}
\textbf{LLM-only methods}
\begin{itemize}[leftmargin=*]
    \item Llama\-2–7B~\cite{touvron2023llama} is a 7B-parameter open-weight model that serves as a mid-scale backbone for instruction-tuned reasoning.
    \item Llama\-3.1–8B ~\cite{dubey2024llama} is a recent 8B variant offering improved alignment and reasoning stability.
    \item Llama\-3.2–3B\protect\footnotemark[1] is a compact 3B model balancing efficiency and compositional reasoning capability.
    \item Llama\-3.2–1B\protect\footnotemark[2] is the smallest model evaluated, used to assess reasoning scalability under extreme parameter constraints.
\end{itemize}
\textbf{KG+LLM Methods}
\begin{itemize}[leftmargin=*]
    \item ToG~\cite{sunthink} explores multiple reasoning paths over KGs and aggregates evidence from them to generate faithful answers, balancing structural exploration and LLM reasoning.
    \item RoG~\cite{luoreasoning} introduces a planning–retrieval–reasoning pipeline where an LLM generates candidate relation sequences that are later grounded in the KG, guiding faithful reasoning.
    \item GCR~\cite{luograph} constrains relation ordering through a KG-Trie mechanism, enforcing structural consistency during path extraction and reducing hallucinations in multi-hop reasoning.
    \item SubgraphRAG~\cite{lisimple} encodes KG triples using a text encoder and trains an MLP classifier to rank and select top-$k$ triplets relevant to the question, enabling structured retrieval without LLM inference.
    \item GNN-RAG~\cite{mavromatis2025gnn} replaces LLM-based retrieval with a lightweight graph neural network that retrieves semantically relevant paths while maintaining efficiency.
\end{itemize}
\subsection{Answer-Centered Prompt Detailed}
\begin{figure}[h]
    \centering
    \includegraphics[width=\linewidth]{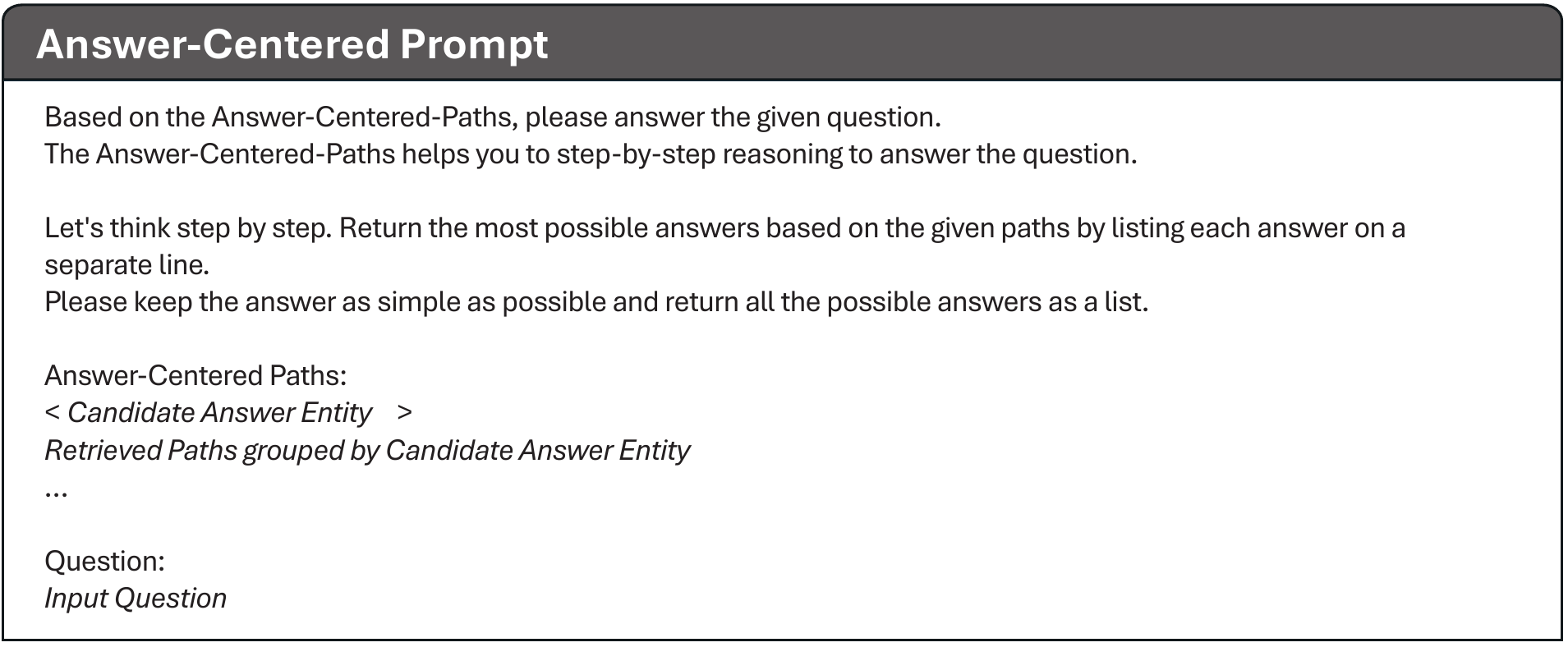}
    \caption{A template of Answer Centered Prompt}
    \label{fig:prompt}
\end{figure}

Figure~\ref{fig:prompt} illustrates the structure of our proposed \textit{answer-centered prompt}. 
In this design, the retrieved results are grouped according to the candidate answer entities, 
allowing the model to focus on reasoning over entity-specific evidence.
\section{Error Case}
\label{adx:error}

\begin{figure}[h]
    \centering
    \includegraphics[width=\linewidth]{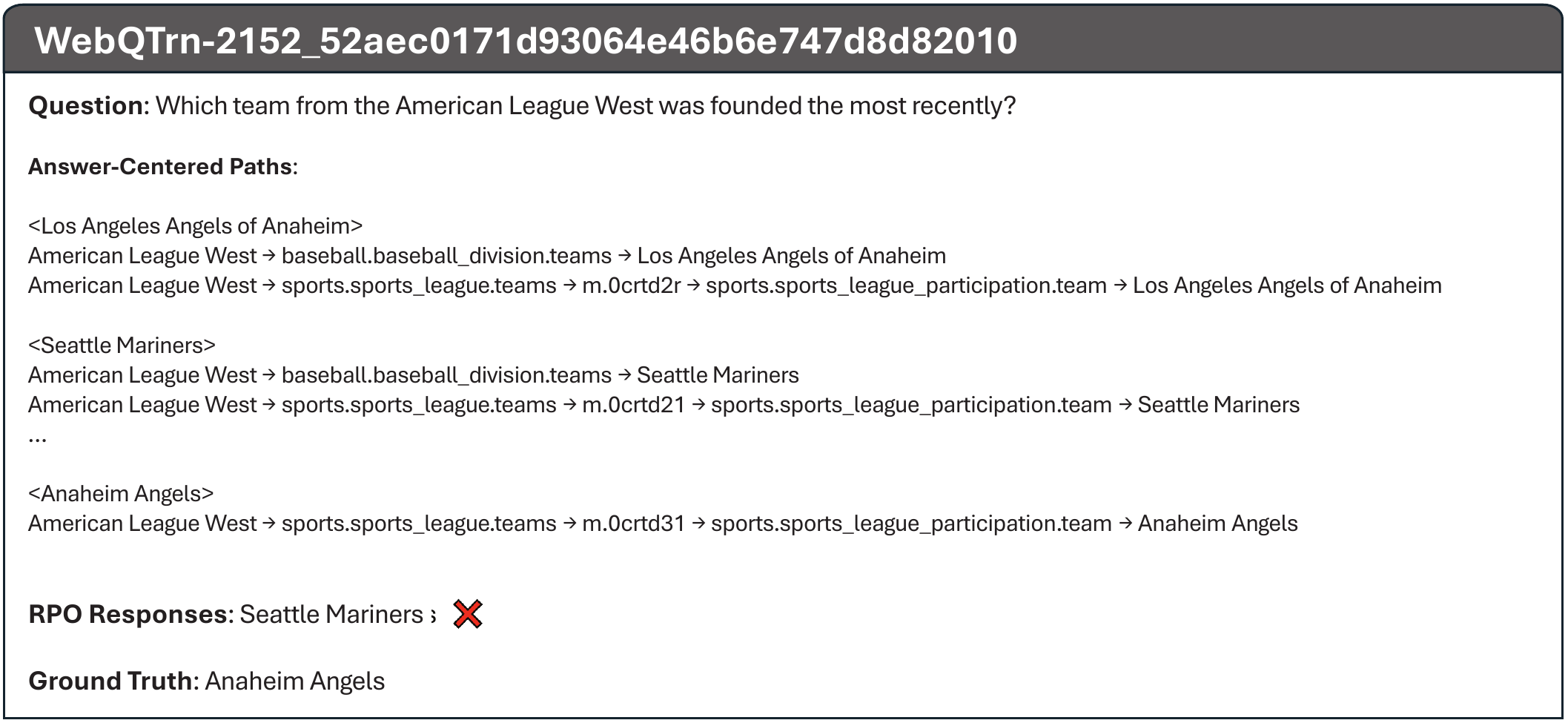}
    \caption{Representative error case of RPO-RAG.  
    The model fails to generate the correct answer despite retrieving valid reasoning paths.}
    \label{fig:error_case}
\end{figure}

In Figure~\ref{fig:error_case}, the example illustrates a failure case where the model does not account for attribute values.
Predicting the correct answer requires reasoning over the attribute values associated with the candidate entities. However, our current approach does not explicitly incorporate such information, leading to an incorrect prediction.
This limitation becomes more pronounced in smaller LLMs ($\leq$3B), where the capacity to infer attribute-dependent relations is further constrained.

\begin{table}[h]
  \centering
  \caption{Detailed hyperparameters for both training stages. LoRA configs are shared across stages.}
  \label{tab:detailed_hparams}
  \setlength{\tabcolsep}{10pt}
  \renewcommand{\arraystretch}{1.05}
  \begin{tabular}{l c}
    \toprule
    \textbf{Hyperparameter} & \textbf{Value} \\
    \midrule
    \multicolumn{2}{l}{\textit{Relation-aware Preference Optimization}} \\
    \cmidrule(lr){1-2}
    \texttt{lora\_r}                   & 32 \\
    \texttt{lora\_alpha}               & 64 \\
    \texttt{lora\_dropout}             & 0.05 \\
    \texttt{optimizer}                 & AdamW \\
    \texttt{warmup\_ratio}             & 0.10 \\
    \texttt{learning\_rate (WebQSP)}   & $7.5\times10^{-6}$ \\
    \texttt{learning\_rate (CWQ)}      & $1.0\times10^{-6}$ \\
    \texttt{scheduler}                 & cosine  \\
    \texttt{max\_length}               & 2048 \\
    \texttt{epochs (WebQSP)}           & 3 \\
    \texttt{epochs (CWQ)}              & 1 \\
    \midrule
    \multicolumn{2}{l}{\textit{answer-centered Prompt Optimization}} \\
    \cmidrule(lr){1-2}
    \texttt{lora\_r}                   & 32 \\
    \texttt{lora\_alpha}               & 64 \\
    \texttt{lora\_dropout}             & 0.05 \\
    \texttt{optimizer}                 & AdamW \\
    \texttt{warmup\_ratio}             & 0.03 \\
    \texttt{learning\_rate}            & $2.0\times10^{-4}$ \\
    \texttt{scheduler}                 & cosine  \\
    \texttt{max\_length}               & 4096 \\
    \texttt{epochs}                    & 3 \\
    \bottomrule
  \end{tabular}
\end{table}

\footnotetext[1]{\url{https://huggingface.co/meta-llama/Llama-3.2-3B-Instruct}}
\footnotetext[2]{\url{https://huggingface.co/meta-llama/Llama-3.2-1B-Instruct}}
\end{document}